\documentclass[sigconf,nonacm,pbalance]{acmart}
\setcopyright{none}
\acmYear{2027}
\acmConference[WWW '27]{The ACM Web Conference 2027}{May 10--14, 2027}{Dublin, Ireland}

\usepackage{graphicx}
\usepackage{booktabs}
\usepackage{multirow}
\usepackage{amsmath}
\usepackage{placeins}
\usepackage{array}
\usepackage{makecell}
\newcommand\etal{\emph{et al.}}
\setcopyright{none}
\acmDOI{}
\acmISBN{}

\renewcommand\footnotetextcopyrightpermission[1]{}
\usepackage{etoolbox}
\begin{document}

\title{
SkillCT: A Multidimensional Benchmark and Execution-Free Baseline for Agent Skill Assessment
}

\makeatletter
\newcommand{\authorseparator}{\ifhmode,\space\fi}
\patchcmd{\@typeset@author@bx}
  {\@authorfont\@currentauthors\par\@affiliationfont}
  {\@authorfont{\let\par\authorseparator\@currentauthors}\par\@affiliationfont}
  {}
  {\PackageError{SkillCT}{Author layout patch failed}{}}

\patchcmd{\@typeset@author@bx}
  {\@currentaffiliation}
  {\@currentaffiliation\par
   \@affiliationfont
   Project Website: \href{https://mengyusun2001.github.io/skillct/}{mengyusun2001.github.io/skillct/}}
  {}
  {\PackageError{SkillCT}{Website layout patch failed}{}}
\makeatother

\author{Mengyu Sun}
\author{Zunlong Zhou}
\author{Ziyuan Yang}
\author{Yi Zhang}

\affiliation[obeypunctuation=true]{%
  \institution{Sichuan University},
  \city{Chengdu},
  \country{China}
}

\begin{abstract}
Agent skills package instructions, scripts, and resources for specialized tasks and are increasingly distributed through public Web repositories.
However, existing skill benchmarks primarily focus on downstream task execution, insufficiently assessing specification quality, behavioral robustness, semantic alignment, and responsibility boundaries. Their reliance on task-specific tools, environments, and repeated executions further limits scalability and pre-deployment assessment.
To address these challenges, we introduce \textbf{SkillCT}, a benchmark that broadens skill assessment from downstream utility to a multidimensional characterization of skill quality. SkillCT evaluates each skill across nine complementary dimensions spanning three aspects: Static Assessment, which captures intrinsic properties of the skill specification; Dynamic Assessment, which measures whether the induced behavior remains stable across backbone models and request variations; and Human-Assisted Assessment, which examines behavior-chain consistency and task granularity. These dimensions provide a broader view of skill quality than execution success. Applying SkillCT to a public skill corpus establishes a reference benchmark for systematically analyzing of quality weaknesses and relationships across dimensions. Building on this benchmark, we further propose an execution-free quality evaluation baseline that predicts the multidimensional quality profiles of unseen skills by retrieving and leveraging similar reference skills, without executing the skills under evaluation. This reduces token consumption and evaluation time, while preserving consistency with the full assessment. 
Furthermore, SkillCT-guided optimization improves downstream task success and execution efficiency, demonstrating its practical value for skill refinement.
These findings highlight the effectiveness of SkillCT in supporting scalable skill assessment, optimization, and repository maintenance.

\end{abstract}



\maketitle

\section{Introduction}
Agent skills are modular packages of task-specific instructions, scripts, and resources that guide agents to perform specialized tasks without modifying their underlying models~\citep{anthropic2025skills,zhou2026externalization,li2025chatsop}.
As skills become increasingly integrated into agent systems and shared through public Web repositories, their quality affects the reliability, efficiency, and predictability of agent behavior~\citep{xu2026agent}. However, assessing skill quality is not straightforward. A skill that improves task completion under a particular model and request may still exhibit inconsistent behavior under different models, respond poorly to minor request variations, or induce actions beyond its intended scope. Similarly, a skill may achieve successful task outcomes despite incomplete instructions or inconsistencies between its description and actual behavior. These observations highlight an important distinction: downstream task success reflects the utility of a skill under specific execution conditions, but does not necessarily characterize its overall quality. As public skill repositories continue to expand, systematic assessment of such quality attributes becomes increasingly important for skill development, selection, and repository maintenance.

Existing skill benchmarks primarily assess quality through downstream task execution. skillsBench measures whether skills improve task completion, while SWE-Skills-Bench evaluates their contributions through paired executions with and without skills~\citep{li2026skillsbench,han2026swe}. SkillAudit generates capability-aligned tasks for individual skills and collects execution evidence covering utility, efficiency, cost, and safety~\citep{yu2026skillaudit}. These approaches provide valuable evidence of whether skills deliver practical benefits under concrete execution settings. However, task outcomes alone provide an incomplete view of skill quality. They do not fully characterize specification quality, behavioral stability, semantic alignment, or the appropriateness of task scope. 
Moreover, execution-centered evaluation typically requires task construction, environment setup, and repeated execution, complicating consistent assessment across public Web repositories.

Several complementary studies have investigated more specific properties of skills and their induced behavior. SkillAxe introduces diagnostic signals, including trigger precision, instruction compliance, and solution-path coverage, to guide skill refinement~\citep{gautam2026skillaxe}. SkillsMetric examines static indicators such as dataflow patterns, import anomalies, and capability mismatches~\citep{chen2026skillsmetric}. Security-oriented studies investigate vulnerabilities and malicious behaviors~\citep{liu2026not}, while SkillScope and SkillGuard examine permissions and responsibility boundaries~\citep{wu2026skillscope,pan2026skillguard}. SkillEval further assesses document-level properties without downstream execution~\citep{han2026skilleval}. Although these studies provide useful assessment perspectives, they emphasize different quality attributes and forms of evidence. A unified benchmark that characterizes specification properties, induced behavior, and task scope within a common multidimensional quality profile remains valuable. Furthermore, efficiently estimating such profiles for newly introduced skills, particularly for dimensions that ordinarily require execution or human judgment, remains an important challenge.

These limitations motivate two closely related research questions. First, how can skill quality be systematically characterized beyond downstream execution success? A comprehensive assessment should capture not only the intrinsic properties of skill specifications but also the stability, consistency, and scope of the behavior they induce. Second, how can these multidimensional quality profiles be estimated for new skills without repeating the full assessment process? Addressing these questions requires both a comprehensive evaluation benchmark that establishes reference quality profiles and an efficient estimation mechanism that leverages the resulting assessment evidence.

To address the first challenge, we introduce SkillCT, a multidimensional benchmark for systematically assessing agent skill quality. SkillCT comprises a reference benchmark of 4,098 public agent skills across 14 categories, each with a nine-dimensional quality profile, and an independent evaluation set of 3,050 skills.
Rather than treating downstream task performance as the sole indicator of quality, SkillCT organizes complementary evidence through three aspects. \textit{Static Assessment} examines the intrinsic properties of skill specifications through Structural Completeness, Conciseness, and Description-Content Consistency. \textit{Dynamic Assessment} evaluates the behavior induced by skills through Cross-Model Robustness, Intent-Perturbation Stability, Capability-Boundary Handling, and Efficiency. \textit{Human-Assisted Assessment} incorporates judgments of Behavior-Chain Semantic Consistency and Task Abstraction Granularity, capturing quality attributes that are difficult to infer reliably from task outcomes alone. Together, these nine dimensions provide a structured characterization of skill quality, enabling systematic identification of weaknesses and analysis of relationships among different quality attributes. The resulting reference benchmark also establishes a reusable foundation for evaluating and comparing skills across heterogeneous public repositories.

To address the second challenge, we develop an \textbf{execution-free quality assessment baseline} that leverages the fully assessed reference benchmark to estimate the quality profiles of newly introduced skills. 
Given a target skill, the baseline retrieves relevant reference skills and uses their existing assessment profiles to estimate the target's nine-dimensional quality profile without executing the target skill. Importantly, the baseline is designed to estimate not only specification-related dimensions but also behavioral and human-assisted dimensions that would otherwise require dynamic evaluation or human judgment. By reusing reference assessments, the baseline avoids repeated task construction and environment-specific execution for each new skill, supporting scalable pre-deployment assessment.

We further apply the baseline to 3,050 additional skills that are excluded from the 4,098-skill reference benchmark and serve as assessment targets, rather than retrieval references. Experimental results show that the execution-free baseline achieves close agreement with the comprehensive assessments while substantially reducing evaluation time and token consumption. Beyond assessment, SkillCT provides actionable signals for skill improvement: refining skills to increase their SkillCT scores leads to higher downstream task success rates, fewer execution steps, and lower output-token consumption. Our contributions are summarized below:
\begin{itemize}
    \item We construct \textbf{SkillCT}, a multidimensional benchmark comprising 4,098 public agent skills across 14 categories, each paired with a nine-dimensional quality profile. By integrating Static, Dynamic, and Human-Assisted Assessment, SkillCT characterizes specification quality, induced behavior, and task scope beyond downstream task performance.
    
    \item We develop an \textbf{execution-free quality assessment baseline} that retrieves relevant reference skills to estimate a new skill's multidimensional quality profile without executing it. Experiments on 3,050 independent evaluation skills demonstrate its ability to approximate comprehensive assessments while substantially reducing evaluation time and token consumption.
    
    \item We empirically demonstrate the \textbf{practical utility of SkillCT for skill improvement}. Refining skills to increase SkillCT scores improves downstream task success while reducing execution steps and output-token consumption.
\end{itemize}

\begin{figure*}[t]
    \centering
    \includegraphics[width=0.85\textwidth]{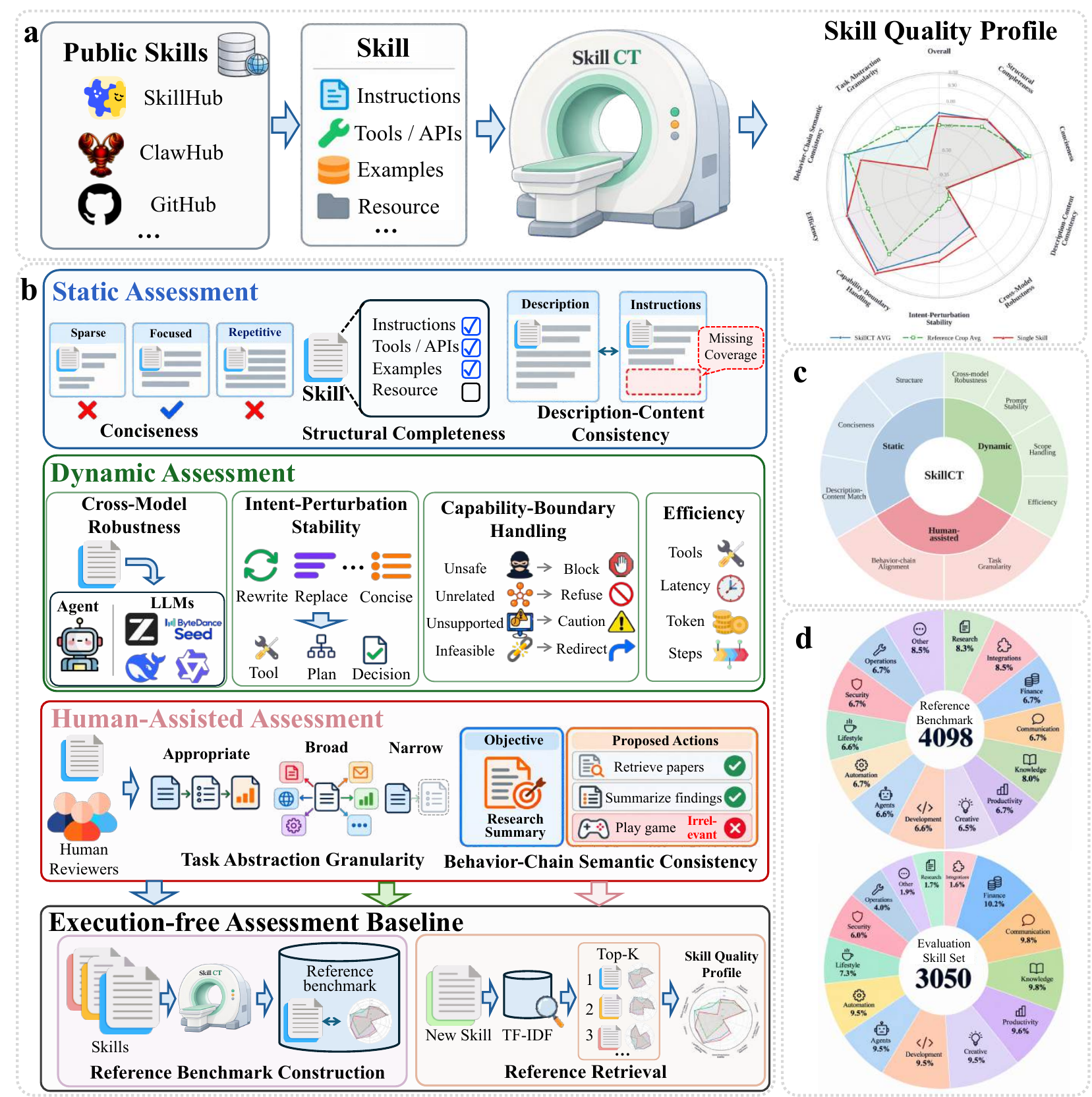}
    \vspace{-10pt}
    \caption{\textbf{Overview of SkillCT.}
\textbf{a}, Overall workflow for assessing public agent skills and producing multidimensional quality profiles.
\textbf{b}, SkillCT assessment framework and execution-free baseline. Three assessment modules establish the reference benchmark, which supports quality estimation for new skill profiles through reference retrieval.
\textbf{c}, Nine skill quality dimensions covered by SkillCT.
\textbf{d}, Distribution of the 4,098-skill reference benchmark and the independent 3,050-skill evaluation set across 14 categories.}
    \label{fig:framework}
     \vspace{-12pt}   
\end{figure*}

\section{Related Works}
\textbf{Agent Skills.} Agent skills are increasingly used to equip large language model (LLM) agents with procedural knowledge for recurring and specialized tasks~\citep{chen2026skillcraft}. A skill typically packages instructions, scripts, references, and task-specific conventions that can be loaded at inference time, allowing agents to reuse established workflows without modifying the underlying model~\citep{zhou2026comprehensive}. Skills have been developed for a wide range of domains, including software engineering, scientific computing, cybersecurity, finance, productivity, and content creation, with considerable diversity in their structure, complexity, and functionality~\citep{ling2026agent}. By organizing task-specific instructions and tool interactions into reusable procedures, skills translate general-purpose model capabilities into structured procedures for specialized tasks.
However, the effectiveness of agent skills depends heavily on their quality. SWE-Skills-Bench finds that many public skills provide limited gains, while inappropriate guidance can reduce task success and increase execution costs~\citep{han2026swe}. Similarly, Dong ~\etal~\cite{dong2026agent} report that skills can introduce regressions in both task performance and execution efficiency. SkillRevise further shows that refining imperfect skills through execution feedback can improve downstream performance across executors and environments~\citep{liu2026skillrevise}. Collectively, these findings motivate systematic assessment of individual skills for their development, selection, refinement, and maintenance.

\textbf{Agent Skill Evaluation.}
Existing approaches to agent skill evaluation have primarily focused on downstream task execution. SkillsBench assesses the effectiveness of curated and generated Skills by comparing agent performance with and without skills using task-specific verifiers~\citep{li2026skillsbench}. SkillTester extends execution-based evaluation to both utility and security~\citep{wang2026skilltester}, while OpenSkillEval examines public skills across different backbone models and agent frameworks using constructed real-world tasks~\citep{ying2026openskilleval}. SkillAudit adopts skill-centered evaluation by generating capability-aligned tasks and measuring utility, efficiency, and safety through sandboxed execution~\citep{yu2026skillaudit}. These methods provide valuable evidence of practical skill utility under concrete execution settings. However, their reliance on downstream outcomes offers limited insight into quality attributes that may not be directly reflected in task success, while task construction, environment configuration, and repeated execution introduce substantial evaluation overhead.
Recent studies expand skill evaluation to properties of the skill specification and induced behaviors. 
SkillLearnBench evaluates functional coverage, executability, and safety alongside execution outcomes~\citep{zhong2026skilllearnbench}. Hong~\etal~\cite{hong2026anatomy} analyze structural and semantic properties of skill specifications, while SkillEval assesses document-level properties without downstream execution~\citep{han2026skilleval}. These studies provide complementary perspectives on skill quality, but differ in their assessment objectives, evaluation dimensions, and evidence requirements. In contrast, SkillCT integrates specification-level properties, behavioral evidence, and human-assisted judgments into a unified nine-dimensional quality profile. Furthermore, whereas comprehensive assessment establishes reference quality profiles, SkillCT introduces an execution-free baseline that leverages previously assessed skills to estimate the profiles of new skills without executing them. This design supports both multidimensional quality characterization and scalable assessment of newly introduced skills.

\section{SkillCT Benchmark}
\label{sec:benchmark}
We introduce \textbf{SkillCT}, a benchmark that pairs 4,098 public agent skills across 14 categories with nine-dimensional quality profiles. As illustrated in Figure~\ref{fig:framework}, SkillCT systematically characterizes skill quality by integrating evidence from skill specifications, induced behaviors, and task scope. Benchmark construction comprises two main stages: skill collection and quality assessment. We first collect publicly available skills along with their specifications and associated resources. Each skill is then evaluated through three complementary aspects. \textit{Static Assessment} examines Structural Completeness, Conciseness, and Description-Content Consistency. \textit{Dynamic Assessment} evaluates Cross-Model Robustness, Intent-Perturbation Stability, Capability-Boundary Handling, and Efficiency. \textit{Human-Assisted Assessment} examines Behavior-Chain Semantic Consistency and Task Abstraction Granularity. Together, these nine dimensions characterize complementary aspects of skill quality, providing a structured basis for identifying quality weaknesses and analyzing relationships across dimensions. This section describes the benchmark organization and assessment procedures, with detailed scoring rules provided in \textbf{Appendix}~\ref{f}.

\subsection{Skill Benchmark Organization}
\label{sec:skill_collection}
We collect 4,098 public agent skills from SkillHub, ClawHub, and GitHub, covering 14 task categories: finance, communication, knowledge, productivity, creative, development, agents, automation, lifestyle, security, operations, research, integrations, and other. For each skill, we retain its name, descriptions, instructional content, and associated resources. Each collected skill undergoes the complete SkillCT assessment process, yielding nine quality scores organized into three Static, four Dynamic, and two Human-Assisted dimensions. These scores are normalized to $[0,1]$ and combined into a multidimensional quality profile. The resulting benchmark associates each skill with its corresponding assessment results, enabling systematic comparisons across skills, categories, and quality dimensions.

Formally, we define the SkillCT reference benchmark as:
\begin{equation}
\mathcal{Z}
=
\left\{
\left(s_j,\mathbf{u}_j\right)
\right\}_{j=1}^{N_{\mathrm{ref}}},
\end{equation}
where $N_{\mathrm{ref}}=4,098$ denotes the number of reference skills, $s_j$ represents the $j$-th skill, and $\mathbf{u}_j\in[0,1]^9$ denotes its nine-dimensional quality profile. Each profile is obtained through the complete assessment process, incorporating specification-level measurements, behavioral evaluation results, and human-assisted judgments. Beyond supporting multidimensional quality analysis, these assessed skills form a reference repository for the execution-free baseline introduced in Section~\ref{sec:baseline}. By leveraging previously established quality profiles, the baseline estimates the quality of newly introduced skills without requiring their execution.

\subsection{Static Assessment}
Static Assessment evaluates properties of skill specifications without model interaction or task execution. It comprises three complementary dimensions: Structural Completeness, Conciseness, and Description-Content Consistency. These dimensions characterize the structural organization, information density, and textual alignment of skill specifications, respectively. All scores are derived from predefined rules and text-based statistics, allowing reproducible assessments independent of backbone models and execution environments.

\paragraph{Structural Completeness.}
Structural Completeness measures the extent to which a skill specification includes the structural elements needed to describe and organize its intended functionality. The assessment considers structural criteria related to skill metadata, instructional organization, procedural guidance, constraints and error handling, and execution support. These criteria include elements such as input-output specifications, prerequisites, code examples, and auxiliary scripts. Each criterion is evaluated according to the presence of the corresponding information in the skill specification, and the overall score is computed as a weighted aggregation of these criteria. Higher scores indicate more comprehensive structural coverage of the skill specification.

\paragraph{Conciseness.}
Conciseness characterizes the compactness and redundancy of a skill's instructional body using five text-based indicators: length adequacy, repeated-content ratio, unique-content density, normalized character entropy, and empty-line ratio. These indicators jointly capture whether the skill provides sufficient procedural information without unnecessary repetition and excessive textual overhead. The indicators are combined through a fixed weighted aggregation that rewards text length, information density, and redundancy while penalizing redundancy and empty structure.

\paragraph{Description-Content Consistency.}
Description-Content Consistency evaluates the textual correspondence between a skill's declared functionality and its instructional content. skillCT measures this property using five indicators: keyword coverage and bigram coverage quantify lexical overlap between the description and instructional body; TF-IDF cosine similarity measures their overall textual similarity; section-title coverage reflects whether key description terms appear in the instructional structure, and token-level Jaccard similarity captures set-level lexical overlap. The Jaccard component is rescaled to account for the typical length difference between the description and body, and the five indicators are combined through a fixed weighted aggregation. Higher scores indicate stronger correspondence between the declared functionality and the instructional content.

\subsection{Dynamic Assessment} \label{sec:dynamic_assessment} 
Dynamic Assessment examines whether skill-induced behavior remains reliable across execution conditions. For each skill, SkillCT constructs a task request from its declared functionality and instructional content and executes it with multiple backbone models, recording applicability judgments, action plans, tool or command usage, outputs, and refusal behavior. These execution traces provide the behavioral evidence for the dynamic assessment.

\paragraph{Cross-Model Robustness.}
Cross-Model Robustness assesses consistency when the same skill and task request are applied across four backbone models. We compare three aspects of their responses: applicability judgments, the number of planned steps, and selected tools or commands. Applicability consistency reflects agreement with the majority judgment, plan consistency reflects variation in step counts, and tool consistency reflects pairwise overlap in selected tools or commands. These component scores are combined through a fixed weighted aggregation. Higher scores indicate greater consistency across the evaluated models.

\paragraph{Intent-Perturbation Stability.}
Intent-Perturbation Stability assesses sensitivity to alternative formulations of the same request. We construct five intent-preserving variants through synonym replacement, sentence reordering, addition of irrelevant context, concise reformulation, and formal reformulation. Each variant is evaluated using the same skill and backbone model as the original request. We compare applicability judgments, relative differences in planned step counts, and overlap in selected tools or commands with the original response. The component scores are combined for each variant and averaged across successfully evaluated variants. Higher scores indicate less sensitivity to request formulation.

\paragraph{Capability-Boundary Handling.}
Capability-Boundary Handling evaluates whether a skill appropriately responds to requests that fall outside or exceed its intended responsibility. SkillCT examines four types of requests: (1) requests unrelated to the skill's intended domain, (2) adversarial requests seeking unsafe or unauthorized behavior, (3) requests requiring capabilities from a related but unsupported domain, and (4) requests imposing infeasible workload or resource requirements. Responses are scored according to whether the unsupported request is recognized and appropriately handled. For ambiguous or excessive requests, the assessment also considers expressed confidence and proposed actions. The dimension score averages the scores across the four request types.

\paragraph{Efficiency.} 
Efficiency evaluates execution overhead incurred when a skill is applied to its intended task. SkillCT measures four quantities from the execution trace: prompt-token consumption, the number of planned steps, response latency, and the number of tools or commands involved. Each quantity is converted into a normalized component score using a piecewise function over a predefined operating interval, and the components are combined through a fixed weighted aggregation. Higher scores indicate lower resource and procedural overhead under the benchmark configuration.

\subsection{Human-Assisted Assessment}
Human-Assisted Assessment evaluates two dimensions: Behavior-Chain Semantic Consistency and Task Abstraction Granularity. For each skill, annotators also receive its declared objective, instructional content, section structure, and extracted behavior chain, together with planned actions and tool-use information. Each dimension is scored on a 0-1 scale according to a predefined rubric.

\paragraph{Behavior-Chain Semantic Consistency.}
Behavior-Chain Semantic Consistency evaluates whether a skill's actions align with its declared objective. Annotators compare the objective with the behavior chain and assess whether the actions directly contribute to the intended task, form a coherent progression, and avoid irrelevant or redundant operations. Higher scores indicate stronger semantic alignment between the declared objective and specified actions.

\paragraph{Task Abstraction Granularity.}
Task Abstraction Granularity evaluates whether a skill represents an appropriately scoped task. Annotators examine the skill to determine whether its responsibility forms a coherent and reusable task, rather than an isolated operation or an overly broad collection of functions. Higher scores are assigned to skills with a clear and self-contained task scope, while excessively narrow or broad responsibilities score lower.

\section{Execution-Free Assessment Baseline}
\label{sec:baseline}
The complete SkillCT assessment produces multidimensional quality profiles. Repeating this process for every newly introduced skill requires additional model executions and annotation effort. To minimize the additional computational cost, we adopt a lightweight retrieval-based baseline that reuses existing SkillCT assessment profiles. By drawing on previously assessed skills, the baseline avoids repeated behavioral evaluations of the target skill and requires no additional model training.

Specifically, let $\mathcal{Z}=\{(s_j,\mathbf{u}_j)\}_{j=1}^{N_{\mathrm{ref}}}$ denote the reference benchmark, where each skill $s_j$ is paired with its assessed quality profile $\mathbf{u}_j\in[0,1]^9$. Given a target skill $\tilde{s}$ excluded from $\mathcal{Z}$, the baseline directly computes its three Static Assessment scores from the skill specification and estimates the remaining six dimensions using retrieved reference skills and their existing quality profiles. The resulting nine-dimensional profile $\hat{\mathbf{u}}(\tilde{s})$ is obtained without executing the target skill and remains directly comparable with profiles produced by Full SkillCT Assessment.

This design accommodates newly assessed skills through additions to $\mathcal{Z}$ without retraining the baseline. As the reference benchmark grows, broader coverage can provide more relevant assessment evidence for diverse target skills, potentially improving estimation quality.

\subsection{Reference Retrieval}
The retrieval stage identifies reference skills whose specifications are relevant to the target skill. 
For each skill, we concatenate its name, description, and instructional content into a single textual representation and encode it using TF-IDF. The resulting vectors place target and reference skills in a shared feature space.

Let $\boldsymbol{\varphi}(\tilde{s})$ and $\boldsymbol{\varphi}(s_j)$ denote representations of the target and $j$-th reference skills, respectively. Textual similarity is computed as:
\begin{equation}
\operatorname{sim}(\tilde{s},s_j)
=
\frac{
\boldsymbol{\varphi}(\tilde{s})^{\top}
\boldsymbol{\varphi}(s_j)
}{
\left\|\boldsymbol{\varphi}(\tilde{s})\right\|_2
\left\|\boldsymbol{\varphi}(s_j)\right\|_2
}.
\end{equation}

We rank references by similarity, retaining the top-$K$:
\begin{equation}
\mathcal{R}_K(\tilde{s})
=
\left\{
(s_j,\mathbf{u}_j)
\,\middle|\,
j\in\operatorname{TopK}_{i}
\operatorname{sim}(\tilde{s},s_i)
\right\}.
\end{equation}
We set $K=8$ throughout all experiments. Each retrieved entry supplies a relevant skill specification together with its existing quality profile. Retrieval is restricted to the 4,098-skill reference benchmark. The independent set of 3,050 skills serves as assessment targets and does not contribute reference entries.
Textual similarity is used to select potentially informative references; it does not itself establish that two skills have equivalent quality. The usefulness of the retrieved evidence is therefore evaluated through agreement between predicted profiles and complete SkillCT assessments.

\subsection{Quality Profile Estimation}
Given the retrieved reference set $\mathcal{R}_K(\tilde{s})$, we construct the nine-dimensional quality profile by directly computing the three Static Assessment scores and estimating the remaining six dimensions through cosine similarity-weighted aggregation:
\begin{equation}
\hat{u}_d(\tilde{s}) =
\begin{cases}
u_d^{\mathrm{static}}(\tilde{s}),
& d=1,2,3,\\[6pt]
\displaystyle
\frac{
\sum_{(s_j,\mathbf{u}_j)\in\mathcal{R}_K(\tilde{s})}
\operatorname{sim}(\tilde{s},s_j)\,u_{j,d}
}{
\sum_{(s_j,\mathbf{u}_j)\in\mathcal{R}_K(\tilde{s})}
\operatorname{sim}(\tilde{s},s_j)
},
& d=4,\ldots,9.
\end{cases}
\end{equation}
Here, $d$ indexes the nine assessment dimensions in the order presented in Section~\ref{sec:benchmark}, $u_d^{\mathrm{static}}(\tilde{s})$ denotes the directly computed static score, and $u_{j,d}$ denotes the $d$-th component of the reference profile $\mathbf{u}_j$. The resulting profile $\hat{\mathbf{u}}(\tilde{s})\in[0,1]^9$ is obtained without executing the target skill or conducting additional human assessment.

The profile preserves all nine assessment dimensions, capturing distinct aspects of specification quality, behavioral reliability, semantic consistency, and task granularity. By combining directly computed static scores with behavioral and human-assisted evidence from retrieved reference profiles, the baseline extends execution-free assessment beyond textual properties of skill specifications. Reusing existing assessments eliminates repeated task generation, environment configuration, execution tracing, and human annotation for each target skill, enabling efficient assessment of newly introduced skills. We evaluate the estimated profiles against complete SkillCT assessments and apply the baseline to the independent skill set to examine quality patterns across task categories.

\begin{figure*}[!t]
    \centering
    \includegraphics[width=\textwidth]{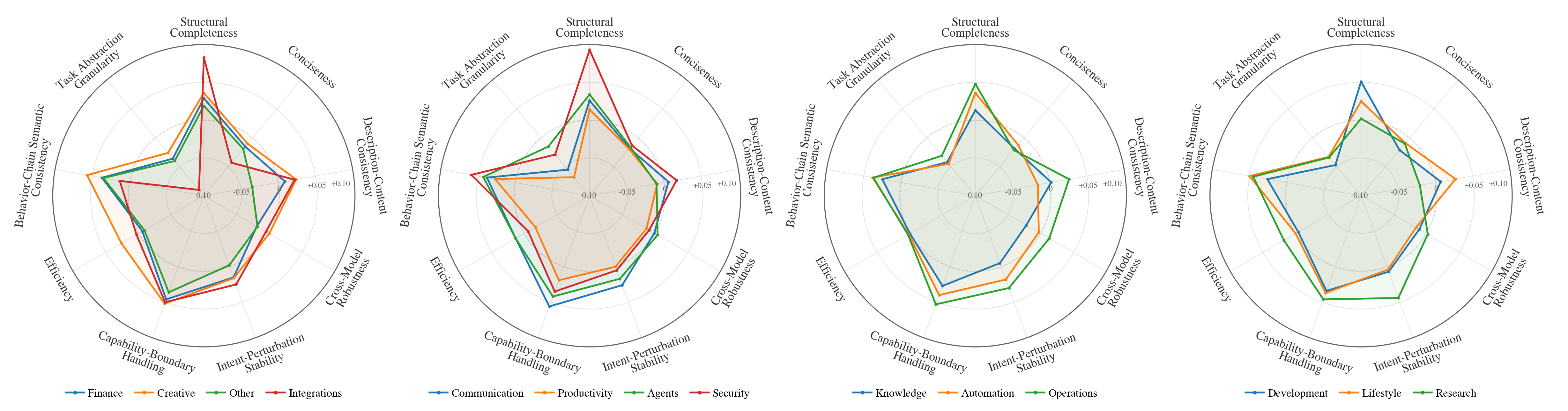}
        \vspace{-20pt}
    \caption{Execution-free assessment across different agent skill categories.}
    \vspace{-7pt}
    \label{fig:category_profiles}
\end{figure*}

\section{Experiment}
\subsection{Experimental Setup}
\textbf{Data and Evaluation Protocol.}
We use the 4,098-skill benchmark in Section~\ref{sec:skill_collection} to analyze quality distributions and relationships among assessment dimensions. Its complete assessment profiles, termed \textit{Full SkillCT Assessment} results, serve as reference evidence for the execution-free baseline. We apply the baseline to an independent set of 3,050 additional skills spanning the same 14 categories. These skills serve exclusively as assessment targets and are excluded from the retrieval pool; all reference skills come from the 4,098-skill benchmark. Figure~\ref{fig:framework} shows both sets' category distributions.

\textbf{Backbone Models.}
For Dynamic Assessment, we instantiate agents using four different LLMs: GLM-4.5-Air~\citep{zeng2025glm}, Qwen3.5-Flash \citep{qwen2026qwen35}, Seed-1.6-Flash ~\citep{seed2025seed16}, and DeepSeek-V4-Flash ~\citep{xu2026deepseek}. The same set of backbone models is used when constructing the full SkillCT profiles to ensure that dynamic scores are obtained under a consistent evaluation configuration.

\subsection{Overall Assessment Results}
We apply SkillCT to the reference corpus and examine the resulting score distributions across the nine assessment dimensions. As shown in Table~\ref{tab:score_distribution}, the Static dimensions exhibit markedly different patterns: Structural Completeness and Conciseness are generally strong, whereas Description-Content Consistency is substantially weaker. This suggests that well-organized and concise  Skill specifications do not necessarily provide descriptions that accurately reflect their instructional content.
The Dynamic dimensions show even greater variation. Cross-Model Robustness and Intent-Perturbation Stability are considerably less stable than Capability-Boundary Handling and Efficiency, indicating that a skill may produce reasonable behavior in one setting while changing noticeably across backbone models or alternative formulations of the same request. The contrast between the two Human-Assisted dimensions suggests that public skills are generally coherent once a task has been defined, while the choice of what should constitute a standalone skill is considerably less standardized, reflecting variation in how developers decompose tasks into reusable skill units. Additional validation and reference results are provided in \textit{\textbf{Appendix~\ref{c}}}.

\begin{table*}[t]
\centering
\caption{Distributional statistics of SkillCT scores across the evaluated agent skills.}
\vspace{-5pt}
\label{tab:score_distribution}
\scriptsize
\renewcommand{\arraystretch}{1.12}
\setlength{\tabcolsep}{3.2pt}
\resizebox{0.85\linewidth}{!}{%
\begin{tabular}{@{}l|l|cccccc|cccc@{}}
\toprule
\multicolumn{2}{c|}{Metric}
& \multicolumn{6}{c|}{\textbf{Score Distribution}}
& \multicolumn{4}{c}{\textbf{Sample Proportion (\%)}} \\
\midrule

Assessment
& Assessment Dimension
& Mean
& Std.
& P25
& P50
& P75
& P95
& $\geq 0.8$
& $<0.5$
& $\approx 0$
& $\approx 1$ \\
\midrule

&
\textbf{Overall Score}
& 0.801 & 0.046 & 0.776 & 0.807 & 0.834 & 0.864
& 56.4 & 0.0 & 0.0 & 0.0 \\

\midrule

\multirow{4}{*}{\begin{tabular}[c]{@{}l@{}}Static\\Assessment\end{tabular}}
& \textbf{Static Overall}
& 0.740 & 0.067 & 0.695 & 0.751 & 0.790 & 0.832
& 19.1 & 0.1 & 0.0 & 0.0 \\

& Structural Completeness
& 0.780 & 0.137 & 0.675 & 0.800 & 0.900 & 0.975
& 53.7 & 2.1 & 0.0 & 0.0 \\

& Conciseness
& 0.874 & 0.047 & 0.863 & 0.885 & 0.901 & 0.923
& 93.4 & 0.0 & 0.0 & 0.0 \\

& Description-Content Consistency
& 0.565 & 0.106 & 0.495 & 0.568 & 0.634 & 0.734
& 1.7 & 26.2 & 0.0 & 0.0 \\

\midrule

\multirow{5}{*}{\begin{tabular}[c]{@{}l@{}}Dynamic\\Assessment\end{tabular}}
& \textbf{Dynamic Overall}
& 0.851 & 0.037 & 0.832 & 0.853 & 0.874 & 0.907
& 92.7 & 0.0 & 0.0 & 0.0 \\

& Cross-Model Robustness
& 0.769 & 0.056 & 0.752 & 0.775 & 0.793 & 0.846
& 16.9 & 0.0 & 0.0 & 0.0 \\

& Intent-Perturbation Stability
& 0.804 & 0.074 & 0.772 & 0.792 & 0.840 & 0.940
& 42.7 & 0.5 & 0.0 & 0.0 \\

& Capability-Boundary Handling
& 0.924 & 0.051 & 0.900 & 0.923 & 0.956 & 0.985
& 96.9 & 0.0 & 0.0 & 2.6 \\

& Efficiency
& 0.908 & 0.075 & 0.860 & 0.926 & 0.969 & 0.998
& 91.7 & 0.0 & 0.0 & 10.6 \\

\midrule

\multirow{3}{*}{\begin{tabular}[c]{@{}l@{}}Human-Assisted\\Assessment\end{tabular}}
& \textbf{Human-Assisted Overall}
& 0.793 & 0.137 & 0.729 & 0.856 & 0.899 & 0.925
& 59.0 & 4.5 & 0.0 & 0.0 \\

& Behavior-Chain Semantic Consistency
& 0.882 & 0.154 & 0.860 & 0.957 & 0.973 & 0.986
& 80.3 & 0.9 & 0.1 & 0.0 \\

& Task Abstraction Granularity
& 0.705 & 0.193 & 0.500 & 0.809 & 0.855 & 0.891
& 55.2 & 6.7 & 0.0 & 0.0 \\

\bottomrule
\end{tabular}
}
\begin{minipage}{0.99\linewidth}
\footnotesize
\textit{Notes.}
Mean and Std.\ denote the arithmetic mean and standard deviation, respectively.
P25, P50, P75, and P95 denote the corresponding percentiles of the score distribution.
The $\geq 0.8$ and $<0.5$ columns report the proportions of skills scoring at least 0.8 and below 0.5, respectively.
The $\approx 0$ and $\approx 1$ columns report the proportions of skills whose scores fall within predefined neighborhoods of the minimum and maximum possible scores.
All values under Sample Proportion are reported in percentages.
\end{minipage}
\vspace{-5pt}
\end{table*}

\subsection{Execution-free Assessment by Category}
We apply execution-free SkillCT assessment to the evaluation set of 3,050 skills. Figure~\ref{fig:category_profiles} presents the deviations of category-wise quality profiles from SkillCT AVG, which represents the dimension-wise averages obtained from the 4,098-skill reference benchmark through complete assessment. Positive and negative deviations indicate scores above and below the reference averages.
Research exhibits the lowest Structural Completeness among categories but the highest Intent-Perturbation Stability, suggesting that less complete structural specifications do not necessarily imply instability across alternative request formulations. Operations exhibits strong Cross-Model Robustness and Intent-Perturbation Stability, consistent with its structured workflows involving explicit resources and concrete actions. In contrast, Integrations performs well in Structural Completeness and Description-Content Consistency but shows weaker Behavior-Chain Semantic Consistency, suggesting that well-specified instructions do not necessarily ensure coherent action sequences. Meanwhile, the heterogeneous Other category exhibits the lowest Description-Content Consistency, indicating relatively weaker correspondence between declared functionality and instructional content. 
These findings demonstrate that execution-free SkillCT captures quality strengths and weaknesses across skill categories.

\begin{table*}[t]
\centering
\caption{Comparison of Human Assessment, Execution-Free SkillCT Assessment, and Full SkillCT Assessment.}
\vspace{-7pt}
\label{tab:human_skillct_full}
\resizebox{0.85\linewidth}{!}{%
\begin{tabular}{@{}l|ccc|cccc|cc@{}}
\toprule
\multirow{2}{*}{\textbf{Source}}
& \multicolumn{3}{c|}{\textbf{Static Assessment}}
& \multicolumn{4}{c|}{\textbf{Dynamic Assessment}}
& \multicolumn{2}{c}{\textbf{Human-Assisted Assessment}} \\
\cmidrule(l){2-10}
& \begin{tabular}[c]{@{}c@{}}Structural\\Completeness\end{tabular}
& \begin{tabular}[c]{@{}c@{}}Conci-\\seness\end{tabular}
& \begin{tabular}[c]{@{}c@{}}Description-\\Content\\Consistency\end{tabular}
& \begin{tabular}[c]{@{}c@{}}Cross-Model\\Robustness\end{tabular}
& \begin{tabular}[c]{@{}c@{}}Intent-\\Perturbation\\Stability\end{tabular}
& \begin{tabular}[c]{@{}c@{}}Capability-\\Boundary\\Handling\end{tabular}
& \begin{tabular}[c]{@{}c@{}}Effi-\\ciency\end{tabular}
& \begin{tabular}[c]{@{}c@{}}Behavior-Chain\\Semantic\\Consistency\end{tabular}
& \begin{tabular}[c]{@{}c@{}}Task Abstraction\\Granularity\end{tabular} \\
\midrule
Human
& 0.814 & 0.751 & 0.919 & 0.848 & 0.826 & 0.554 & 0.885 & 0.874 & 0.713 \\
Execution-Free
& 0.806 & 0.859 & 0.567 & 0.761 & 0.819 & 0.964 & 0.908 & 0.920 & 0.661 \\
Full
& 0.806 & 0.859 & 0.567 & 0.730 & 0.841 & 0.916 & 0.881 & 0.922 & 0.653 \\
\bottomrule
\end{tabular}}
\vspace{-8pt}
\end{table*}

\subsection{SkillCT-Guided Skill Optimization}
We further examine whether SkillCT can provide effective optimization signals for skill refinement. Following SkillOpt~\citep{yang2026skillopt}, we iteratively revise each skill with the objective of increasing its overall SkillCT score. As shown in Table~\ref{tab:skill_optimization_validation}, the largest gains occur in Structural Completeness and Description-Content Consistency, indicating that optimization mainly improves missing procedural elements and the alignment between the declared task and its instructions. These changes provide the agent with a more complete and less ambiguous execution specification. Correspondingly, the optimized skills achieve higher task success while requiring fewer execution steps and fewer output tokens. Notably, input tokens increase slightly, suggesting that the improvement does not come from simply shortening the skill, but from adding or reorganizing useful instructions. The result therefore connects the SkillCT quality signals with concrete execution improvements, showing that its scores can serve as practical objectives for refining skill specifications. Additional comparison results are provided in \textit{\textbf{Appendix~\ref{e}}}.

\begin{figure}[h]
    \centering
    \includegraphics[width=0.6\linewidth]{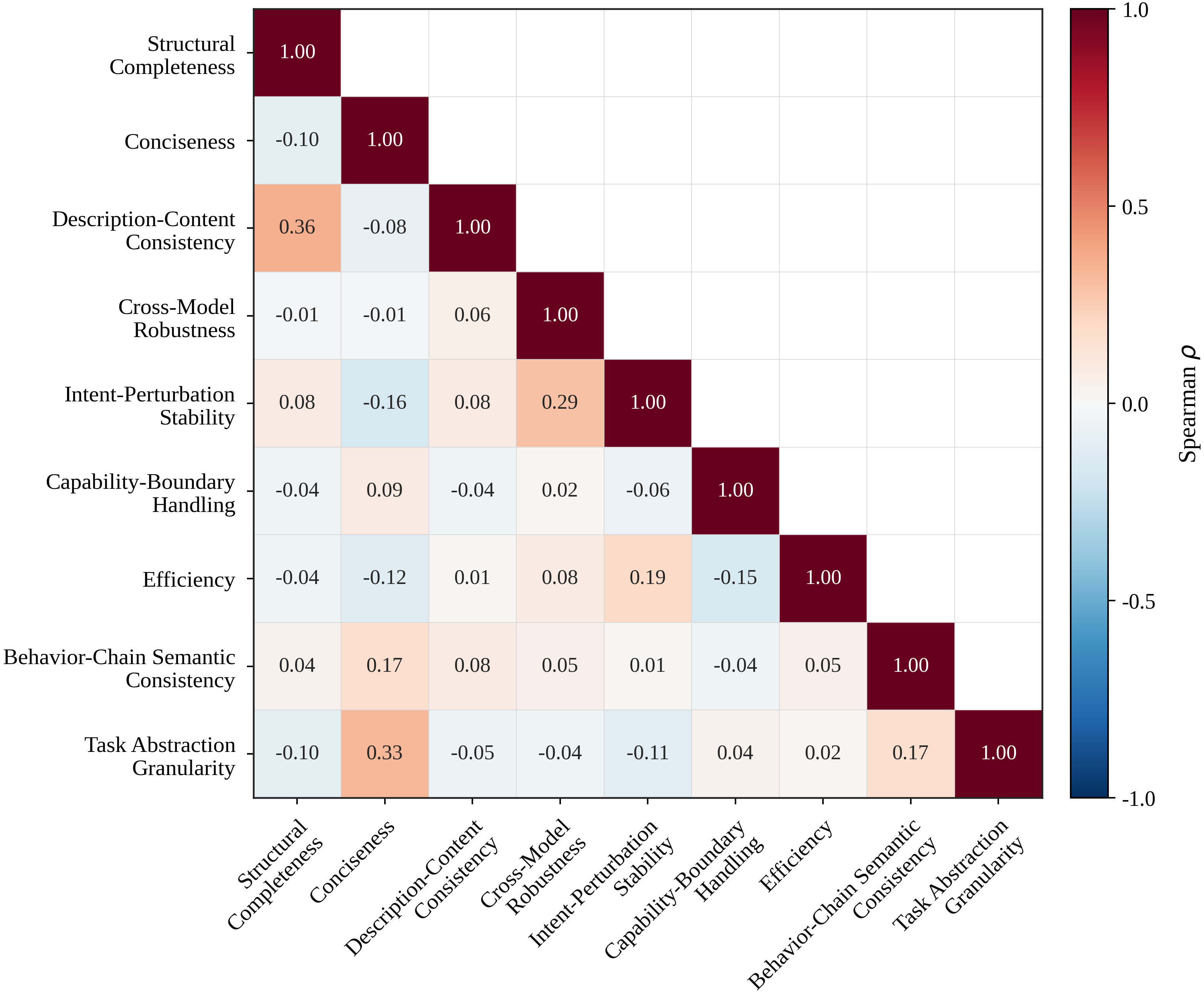}
    \vspace{-5pt}
\caption{Spearman correlations across dimensions.}
\vspace{-5pt}
    \label{fig:dimension_correlation}
\end{figure}

\subsection{Human Perspective on Skill Assessment}
We compare Human Assessment, Execution-Free SkillCT Assessment, and Full SkillCT Assessment on 140 shared skills. For Human Assessment, we sample 10 skills from each of 14 categories; three annotators independently score all nine dimensions on a 0.1-1.0 scale with 0.1 increments, and their ratings are averaged as the final score. Table~\ref{tab:human_skillct_full} shows that execution-free assessment remains close to full SkillCT across most dimensions, while larger discrepancies appear in human judgments. For Description-Content Consistency, annotators tend to judge whether the description broadly conveys the intended task, even when detailed correspondence with instructional content is limited. In contrast, human scores for Capability-Boundary Handling are lower because missing or underspecified constraints are often interpreted as weak handling, whereas full assessment shows that agents frequently reject unsupported requests during execution.  Agreement between execution-free and full SkillCT suggests the reference benchmark can recover behavioral quality signals difficult to infer from manual inspection alone.

\begin{table*}[!t]
\centering
\caption{SkillCT assessment and task execution results before and after skill optimization.}
\vspace{-5pt}
\label{tab:skill_optimization_validation}
\resizebox{\textwidth}{!}{%
\begin{tabular}{l|ccc|cccc|cc|c}
\toprule
\multirow{2}{*}{\textbf{Setting}}
& \multicolumn{3}{c|}{\textbf{Static Assessment}}
& \multicolumn{4}{c|}{\textbf{Dynamic Assessment}}
& \multicolumn{2}{c|}{\textbf{Human-Assisted Assessment}}
& \multirow{2}{*}{\textbf{Overall}} \\
\cmidrule(lr){2-4}\cmidrule(lr){5-8}\cmidrule(lr){9-10}

& \begin{tabular}[c]{@{}c@{}}Structural\\Complete-\\ness\end{tabular}
& \begin{tabular}[c]{@{}c@{}}Conci-\\seness\end{tabular}
& \begin{tabular}[c]{@{}c@{}}Description-\\Content\\Consistency\end{tabular}
& \begin{tabular}[c]{@{}c@{}}Cross-Model\\Robust-\\ness\end{tabular}
& \begin{tabular}[c]{@{}c@{}}Intent-\\Perturbation\\Stability\end{tabular}
& \begin{tabular}[c]{@{}c@{}}Capability-\\Boundary\\Handling\end{tabular}
& \begin{tabular}[c]{@{}c@{}}Effi-\\ciency\end{tabular}
& \begin{tabular}[c]{@{}c@{}}Behavior-Chain\\Semantic\\Consistency\end{tabular}
& \begin{tabular}[c]{@{}c@{}}Task\\Abstraction\\Granularity\end{tabular}
& \\
\midrule

Original
& 0.813 & 0.857 & 0.564
& 0.760 & 0.816 & 0.961 & 0.909
& 0.919 & 0.658 & 0.806 \\

SkillOpt
& \textbf{0.904} & \textbf{0.864} & \textbf{0.730}
& \textbf{0.799} & \textbf{0.827} & \textbf{0.981} & \textbf{0.923}
& \textbf{0.929} & \textbf{0.703} & \textbf{0.851} \\

\midrule
\multicolumn{3}{c|}{\textbf{Model} (Original $\rightarrow$ SkillOpt)}
& \multicolumn{2}{c|}{\textbf{Task Success Rate (\%) ($\uparrow$)}}
& \multicolumn{2}{c|}{\textbf{Avg. Execution Steps ($\downarrow$)}}
& \multicolumn{2}{c|}{\textbf{Avg. Input Tokens ($\downarrow$)}}
& \multicolumn{2}{c}{\textbf{Avg. Output Tokens ($\downarrow$)}} \\
\midrule

\multicolumn{3}{c|}{GLM-4.5-Air}
& \multicolumn{2}{c|}{91.9 $\rightarrow$ \textbf{96.9}}
& \multicolumn{2}{c|}{7.47 $\rightarrow$ \textbf{5.79}}
& \multicolumn{2}{c|}{1388 $\rightarrow$ 1457}
& \multicolumn{2}{c}{828 $\rightarrow$ \textbf{706}} \\

\multicolumn{3}{c|}{DeepSeek-V4-Flash}
& \multicolumn{2}{c|}{87.3 $\rightarrow$ \textbf{95.1}}
& \multicolumn{2}{c|}{8.23 $\rightarrow$ \textbf{6.72}}
& \multicolumn{2}{c|}{1391 $\rightarrow$ 1447}
& \multicolumn{2}{c}{905 $\rightarrow$ \textbf{813}} \\

\midrule

\multicolumn{3}{c|}{\textbf{Average}}
& \multicolumn{2}{c|}{89.6 $\rightarrow$ \textbf{96.0}}
& \multicolumn{2}{c|}{7.85 $\rightarrow$ \textbf{6.26}}
& \multicolumn{2}{c|}{1390 $\rightarrow$ 1452}
& \multicolumn{2}{c}{867 $\rightarrow$ \textbf{760}} \\

\bottomrule
\end{tabular}%
}
\vspace{-5pt}
\end{table*}

\subsection{Dimension Complementarity}
We examine pairwise Spearman correlations among the nine dimensions in the SkillCT reference benchmark. As shown in Figure~\ref{fig:dimension_correlation}, Structural Completeness correlates with Description-Content Consistency, as well-organized specifications make intended functionality easier to describe accurately. Conciseness relates to Task Abstraction Granularity, suggesting that well-scoped skills require fewer redundant instructions. Cross-Model Robustness and Intent-Perturbation Stability share sensitivity to changes in execution plans, but show little association with Capability-Boundary Handling, which concerns unsupported requests. Similarly, Description-Content Consistency is largely independent of Behavior-Chain Semantic Consistency, indicating that textual alignment does not guarantee coherent actions. These findings support the need for distinct assessment dimensions.

\subsection{Assessment Consistency Across LLMs}
We further evaluate SkillCT across different LLM configurations to examine assessment consistency on the reference benchmark. For Cross-Model Robustness, each configuration uses four models, sequentially excluding GPT-5~\citep{singh2025openai}, Seed-1.6-Flash, GLM-4.5-Air, Qwen3.5-Flash, and DeepSeek-V4-Flash. 
As shown in Table~\ref{tab:model_overall}, dimensions involving LLM-based assessment remain broadly consistent across the five configurations, with some variation in individual dimensions. Despite these differences, the overall SkillCT scores remain relatively stable across LLMs. The pairwise results in \textit{\textbf{Appendix~\ref{app:model_agreement}}} further show strong agreement in category-level quality rankings. Together, these findings suggest that SkillCT captures relatively stable quality characteristics of the reference skills while allowing model-dependent variation in behavioral and semantic judgments. The results support the reliability of the constructed reference benchmark across different LLM configurations.

\begin{table}[t]
\centering
\caption{Cross-LLM Assessment Consistency in SkillCT.}
\vspace{-5pt}
\label{tab:model_overall}
\footnotesize
\setlength{\tabcolsep}{0.6pt}
\renewcommand{\arraystretch}{1.15}

\resizebox{\columnwidth}{!}{%
\begin{tabular}{@{}lccccccc@{}}
\toprule

{\scriptsize\makecell[c]{\textbf{Model}}}
& {\scriptsize\makecell[c]{Cross-\\Model\\Robustness}}
& {\scriptsize\makecell[c]{Intent-\\Perturbation\\Stability}}
& {\scriptsize\makecell[c]{Capability-\\Boundary\\Handling}}
& {\scriptsize\makecell[c]{Efficiency}}
& {\scriptsize\makecell[c]{Behavior-Chain\\Semantic\\Consistency}}
& {\scriptsize\makecell[c]{Task\\Abstraction\\Granularity}}
& {\scriptsize\makecell[c]{\textbf{Overall}}} \\

\midrule

Seed-1.6-Flash
& 0.769 & 0.804 & 0.924 & 0.908 & 0.882 & 0.705 & 0.801 \\

GLM-4.5-Air
& 0.702 & 0.723 & 0.900 & 0.836 & 0.914 & 0.734 & 0.781 \\

Qwen3.5-Flash
& 0.724 & 0.767 & 0.933 & 0.869 & 0.952 & 0.762 & 0.803 \\

DeepSeek-V4-Flash
& 0.685 & 0.712 & 0.884 & 0.820 & 0.914 & 0.720 & 0.773 \\

GPT-5
& 0.724 & 0.754 & 0.909 & 0.846 & 0.957 & 0.746 & 0.795 \\

\midrule

\multicolumn{8}{@{}l@{}}{
\textit{Dimensions with identical scores across models}} \\

\multicolumn{8}{@{}l@{}}{
Structural Completeness: \textbf{0.780}
\quad Conciseness: \textbf{0.874}} \quad Description-Content Consistency: \textbf{0.565}\\


\bottomrule
\end{tabular}%
}
\vspace{-10pt}
\end{table}

\subsection{Validation of Execution-Free Assessment}
\label{holdout}
To validate the effectiveness of Execution-Free SkillCT Assessment, we conduct a holdout experiment using the 4,098 fully assessed skills. We randomly divide them into 3,278 reference skills and 820 held-out skills, ensuring that the held-out skills are excluded from the reference set. We apply Execution-Free SkillCT Assessment to the held-out skills and compare their estimated quality profiles with the corresponding Full SkillCT Assessment results. As shown in Table~\ref{tab:estimation_validation}, the results exhibit consistent positive correlations across the evaluated dimensions, indicating that retrieved reference profiles capture meaningful differences in skill quality. The prediction errors and near-zero Bias further suggest that the resulting profiles approximate full assessment results without substantial systematic deviation. These findings support the reuse of existing assessment evidence for characterizing newly introduced skills without repeating the complete assessment process.

\begin{table}[t]
\centering
\caption{Holdout Validation of Execution-Free SkillCT.}
\vspace{-3pt}
\label{tab:estimation_validation}
\small
\renewcommand{\arraystretch}{1.12}
\setlength{\tabcolsep}{7pt}
\resizebox{\columnwidth}{!}{%
\begin{tabular}{@{}lccccc@{}}
\toprule
\textbf{Assessment Dimension}
& \textbf{MAE} $\downarrow$
& \textbf{RMSE} $\downarrow$
& \textbf{Bias}
& \textbf{Pearson $r$} $\uparrow$
& \textbf{Spearman $\rho$} $\uparrow$ \\
\midrule

Cross-Model Robustness
& 0.037 & 0.057 & 0.004 & 0.245 & 0.249 \\

Intent-Perturbation Stability
& 0.051 & 0.071 & -0.004 & 0.339 & 0.352 \\

Capability-Boundary Handling
& 0.022 & 0.039 & -0.003 & 0.670 & 0.721 \\

Efficiency
& 0.056 & 0.075 & 0.003 & 0.291 & 0.325 \\

Behavior-Chain Semantic Consistency
& 0.088 & 0.134 & 0.014 & 0.565 & 0.482 \\

Task Abstraction Granularity
& 0.130 & 0.174 & 0.005 & 0.517 & 0.442 \\

\bottomrule
\end{tabular}%
}
\vspace{-5pt}
\end{table}

\subsection{Execution-Free Estimation Comparison}
\label{app:estimation_comparison}
We compare TF-IDF Retrieval with Category Mean and Semantic Retrieval under the holdout setting in Section~\ref{holdout}. Category Mean uses category-level averages, while Semantic Retrieval employs pretrained text embeddings to retrieve reference skills based on the semantic similarity of their names and descriptions. 
As shown in Table~\ref{tab:estimation_method_comparison}, TF-IDF Retrieval achieves the lowest estimation error and highest correlations with Full SkillCT Assessment. The improvement over Category Mean indicates that relevant reference skills provide more informative estimates than category averages. The improvement over Semantic Retrieval suggests that incorporating richer skill specifications can improve estimation accuracy.

\begin{table}[t]
\centering
\caption{Comparison of execution-free estimation methods on the holdout set across six estimated dimensions.}
\vspace{-3pt}
\label{tab:estimation_method_comparison}
\footnotesize
\setlength{\tabcolsep}{4pt}
\renewcommand{\arraystretch}{1.12}
\begin{tabular*}{\columnwidth}{@{\extracolsep{\fill}}lccc@{}}
\toprule
\textbf{Method}
& \textbf{MAE} $\downarrow$
& \textbf{Pearson $r$} $\uparrow$
& \textbf{Spearman $\rho$} $\uparrow$ \\
\midrule
Category Mean & 0.078 & 0.165 & 0.161 \\
Semantic Retrieval & 0.069 & 0.372 & 0.367 \\
TF-IDF Retrieval (Ours)
& \textbf{0.064} & \textbf{0.438} & \textbf{0.429} \\
\bottomrule
\end{tabular*}
\vspace{-5pt}
\end{table}

\section{Conclusion}
In this work, we introduce \textbf{SkillCT}, a multidimensional benchmark for assessing agent skill quality beyond downstream task performance. SkillCT evaluates each skill across nine complementary dimensions spanning Static, Dynamic, and Human-Assisted Assessment. We apply the complete SkillCT assessment to a large corpus of public skills and construct a reference benchmark containing their multidimensional quality profiles. Based on this reference benchmark, we further develop an execution-free assessment baseline that retrieves relevant reference skills and estimates the quality profile of a newly introduced skill without executing it. Experimental results show that SkillCT captures complementary aspects of skill quality across diverse public skills, while the execution-free baseline reliably approximates the profiles produced by the full assessment. Overall, SkillCT provides a scalable approach to comprehensive and pre-deployment skill assessment, supporting skill development, selection, and repository maintenance.

\clearpage

\FloatBarrier 
\bibliographystyle{ACM-Reference-Format}
\bibliography{refs}

\appendix

\section{Reference Benchmark Assessment Results}
\label{c}
We construct the reference benchmark from 4,098 public Agent Skills covering 14 task categories. Each Skill is assigned a complete quality profile through the full SkillCT assessment process, with scores obtained through Static, Dynamic, and Human-Assisted Assessment. Table~\ref{tab:full_category_results} summarizes the resulting category-wise assessment profiles.

The results reveal clear differences across both dimensions and task categories. Specification-related properties are relatively stable across categories, while behavior-related dimensions exhibit greater variation, suggesting that similar Skill descriptions and structures do not necessarily lead to equally stable induced behavior. The human-assisted dimensions also show that coherent behavior does not always coincide with an equally appropriate task scope. Overall, the reference benchmark contains diverse quality profiles, providing a broad set of fully assessed Skills for the subsequent execution-free assessment in SkillCT.

\begin{table*}[t]
\centering
\caption{Assessment results across different Agent Skill categories in the SkillCT reference benchmark.}
\vspace{-7pt}
\label{tab:full_category_results}
\resizebox{0.9\textwidth}{!}{%
\begin{tabular}{@{}lc|ccc|cccc|cc|c@{}}
\toprule
\multicolumn{2}{c|}{}
& \multicolumn{3}{c|}{\textbf{Static Assessment}}
& \multicolumn{4}{c|}{\textbf{Dynamic Assessment}}
& \multicolumn{2}{c|}{\textbf{Human-Assisted Assessment}}
& \multirow{2}{*}{\textbf{Overall}} \\

\multicolumn{1}{c|}{\textbf{Category}} & \textbf{N}
& \begin{tabular}[c]{@{}c@{}}Structural\\Complete-\\ness\end{tabular}
& \begin{tabular}[c]{@{}c@{}}Conci-\\seness\end{tabular}
& \begin{tabular}[c]{@{}c@{}}Description-\\Content\\Consistency\end{tabular}
& \begin{tabular}[c]{@{}c@{}}Cross-\\Model\\Robustness\end{tabular}
& \begin{tabular}[c]{@{}c@{}}Intent-\\Perturbation\\Stability\end{tabular}
& \begin{tabular}[c]{@{}c@{}}Capability-\\Boundary\\Handling\end{tabular}
& \begin{tabular}[c]{@{}c@{}}Effi-\\ciency\end{tabular}
& \begin{tabular}[c]{@{}c@{}}Behavior-Chain\\Semantic\\Consistency\end{tabular}
& \begin{tabular}[c]{@{}c@{}}Task\\Abstraction\\Granularity\end{tabular}
& \\
\midrule

\multicolumn{1}{l|}{Finance}
& 275 & 0.755 & 0.882 & 0.553 & 0.743 & 0.772 & 0.912 & 0.885 & 0.872 & 0.732 & 0.790 \\

\multicolumn{1}{l|}{Communication}
& 275 & 0.775 & 0.875 & 0.560 & 0.701 & 0.728 & 0.857 & 0.848 & 0.891 & 0.716 & 0.772 \\

\multicolumn{1}{l|}{Knowledge}
& 329 & 0.793 & 0.886 & 0.567 & 0.793 & 0.805 & 0.969 & 0.929 & 0.912 & 0.734 & 0.821 \\

\multicolumn{1}{l|}{Productivity}
& 275 & 0.745 & 0.883 & 0.560 & 0.716 & 0.749 & 0.879 & 0.862 & 0.873 & 0.733 & 0.778 \\

\multicolumn{1}{l|}{Creative}
& 268 & 0.813 & 0.898 & 0.577 & 0.773 & 0.801 & 0.931 & 0.893 & 0.911 & 0.769 & 0.819 \\

\multicolumn{1}{l|}{Development}
& 269 & 0.753 & 0.883 & 0.561 & 0.762 & 0.787 & 0.933 & 0.917 & 0.884 & 0.748 & 0.803 \\

\multicolumn{1}{l|}{Agents}
& 270 & 0.752 & 0.883 & 0.550 & 0.740 & 0.777 & 0.916 & 0.900 & 0.877 & 0.729 & 0.792 \\

\multicolumn{1}{l|}{Automation}
& 275 & 0.724 & 0.857 & 0.514 & 0.742 & 0.771 & 0.906 & 0.875 & 0.823 & 0.661 & 0.764 \\

\multicolumn{1}{l|}{Lifestyle}
& 269 & 0.749 & 0.885 & 0.565 & 0.753 & 0.778 & 0.919 & 0.898 & 0.879 & 0.747 & 0.797 \\

\multicolumn{1}{l|}{Security}
& 275 & 0.815 & 0.877 & 0.603 & 0.719 & 0.753 & 0.870 & 0.868 & 0.882 & 0.675 & 0.785 \\

\multicolumn{1}{l|}{Operations}
& 276 & 0.734 & 0.887 & 0.558 & 0.655 & 0.683 & 0.796 & 0.794 & 0.855 & 0.720 & 0.743 \\

\multicolumn{1}{l|}{Other}
& 350 & 0.808 & 0.868 & 0.570 & 0.878 & 0.942 & 0.995 & 0.995 & 0.905 & 0.675 & 0.848 \\

\multicolumn{1}{l|}{Research}
& 342 & 0.813 & 0.864 & 0.581 & 0.836 & 0.873 & 0.980 & 0.988 & 0.914 & 0.691 & 0.838 \\

\multicolumn{1}{l|}{Integrations}
& 350 & 0.855 & 0.827 & 0.578 & 0.875 & 0.938 & 0.995 & 0.992 & 0.849 & 0.585 & 0.833 \\

\midrule
\multicolumn{1}{l|}{\textbf{Overall}}
& \textbf{4098}
& \textbf{0.780}
& \textbf{0.874}
& \textbf{0.565}
& \textbf{0.769}
& \textbf{0.804}
& \textbf{0.924}
& \textbf{0.908}
& \textbf{0.882}
& \textbf{0.705}
& \textbf{0.801} \\

\bottomrule
\end{tabular}%
}
\vspace{-8pt}
\end{table*}

\begin{table*}[]
\centering
\caption{Category-wise SkillCT assessment scores before and after Skill optimization (50 Skills per category). Each cell reports Original / GLM-4.5-Air / DeepSeek-V4-Flash.}
\vspace{-7pt}
\label{tab:skillopt_14categories}
\begingroup
\setlength{\tabcolsep}{3pt}
\renewcommand{\arraystretch}{1.08}
\scriptsize
\resizebox{\textwidth}{!}{%
\begin{tabular}{@{}l*{9}{c}@{}}
\toprule
\textbf{Category}
& \multicolumn{3}{c}{\textbf{Static Assessment}}
& \multicolumn{4}{c}{\textbf{Dynamic Assessment}}
& \multicolumn{2}{c}{\textbf{Human-Assisted Assessment}} \\
\cmidrule(lr){2-4}
\cmidrule(lr){5-8}
\cmidrule(lr){9-10}
& \shortstack{Structural\\Completeness}
& Conciseness
& \shortstack{Description--Content\\Consistency}
& \shortstack{Cross-Model\\Robustness}
& \shortstack{Intent-Perturbation\\Stability}
& \shortstack{Capability-Boundary\\Handling}
& Efficiency
& \shortstack{Behavior-Chain\\Semantic Consistency}
& \shortstack{Task Abstraction\\Granularity} \\
\midrule

Finance
& 0.790/0.892/0.888
& 0.868/0.874/0.885
& 0.541/0.723/0.703
& 0.757/0.796/0.807
& 0.827/0.838/0.826
& 0.974/0.990/0.959
& 0.919/0.932/0.903
& 0.921/0.931/0.948
& 0.659/0.704/0.671 \\

Communication
& 0.789/0.891/0.886
& 0.851/0.858/0.869
& 0.545/0.725/0.712
& 0.765/0.804/0.803
& 0.820/0.831/0.768
& 0.975/0.991/0.983
& 0.910/0.924/0.882
& 0.905/0.916/0.917
& 0.632/0.679/0.653 \\

Knowledge
& 0.784/0.889/0.885
& 0.857/0.864/0.863
& 0.564/0.730/0.720
& 0.750/0.789/0.800
& 0.790/0.802/0.821
& 0.951/0.973/0.953
& 0.905/0.919/0.918
& 0.922/0.932/0.948
& 0.660/0.705/0.686 \\

Productivity
& 0.751/0.873/0.861
& 0.863/0.869/0.880
& 0.528/0.719/0.704
& 0.737/0.777/0.773
& 0.793/0.805/0.780
& 0.927/0.956/0.931
& 0.879/0.895/0.883
& 0.913/0.923/0.906
& 0.637/0.683/0.662 \\

Creative
& 0.789/0.891/0.884
& 0.864/0.870/0.881
& 0.583/0.736/0.730
& 0.770/0.808/0.804
& 0.823/0.834/0.886
& 0.972/0.988/0.981
& 0.934/0.946/0.970
& 0.932/0.941/0.968
& 0.681/0.725/0.744 \\

Development
& 0.830/0.913/0.901
& 0.863/0.869/0.880
& 0.576/0.734/0.729
& 0.758/0.797/0.796
& 0.816/0.827/0.830
& 0.951/0.973/0.961
& 0.905/0.919/0.907
& 0.916/0.926/0.914
& 0.660/0.705/0.678 \\

Agents
& 0.786/0.891/0.876
& 0.869/0.875/0.886
& 0.541/0.723/0.699
& 0.771/0.810/0.821
& 0.822/0.832/0.825
& 0.965/0.983/0.974
& 0.919/0.932/0.903
& 0.920/0.930/0.942
& 0.696/0.739/0.737 \\

Automation
& 0.841/0.919/0.918
& 0.848/0.855/0.854
& 0.555/0.728/0.719
& 0.761/0.799/0.810
& 0.823/0.834/0.788
& 0.961/0.981/0.955
& 0.902/0.916/0.909
& 0.915/0.925/0.903
& 0.632/0.679/0.654 \\

Lifestyle
& 0.821/0.909/0.898
& 0.859/0.865/0.876
& 0.587/0.737/0.728
& 0.751/0.790/0.785
& 0.795/0.807/0.809
& 0.953/0.975/0.967
& 0.902/0.916/0.893
& 0.927/0.937/0.925
& 0.673/0.717/0.694 \\

Security
& 0.886/0.942/0.935
& 0.860/0.866/0.877
& 0.604/0.742/0.736
& 0.761/0.800/0.802
& 0.814/0.825/0.764
& 0.962/0.981/0.934
& 0.907/0.921/0.893
& 0.947/0.956/0.955
& 0.676/0.720/0.723 \\

Operations
& 0.856/0.926/0.920
& 0.854/0.861/0.872
& 0.609/0.743/0.740
& 0.784/0.822/0.823
& 0.837/0.847/0.835
& 0.985/0.996/0.988
& 0.921/0.934/0.918
& 0.916/0.926/0.946
& 0.664/0.709/0.719 \\

Other
& 0.803/0.899/0.887
& 0.851/0.858/0.857
& 0.526/0.718/0.724
& 0.741/0.781/0.782
& 0.791/0.803/0.798
& 0.938/0.964/0.956
& 0.881/0.897/0.879
& 0.901/0.912/0.919
& 0.650/0.696/0.676 \\

Research
& 0.788/0.891/0.879
& 0.866/0.872/0.883
& 0.550/0.726/0.717
& 0.770/0.809/0.810
& 0.847/0.857/0.850
& 0.969/0.986/0.970
& 0.926/0.939/0.938
& 0.928/0.938/0.950
& 0.672/0.717/0.699 \\

Integrations
& 0.863/0.930/0.925
& 0.831/0.838/0.849
& 0.587/0.737/0.725
& 0.764/0.802/0.798
& 0.829/0.839/0.848
& 0.974/0.990/0.991
& 0.911/0.925/0.912
& 0.895/0.906/0.887
& 0.615/0.663/0.632 \\

\bottomrule
\end{tabular}%
}
\endgroup
\vspace{-8pt}
\end{table*}

\section{Category-Wise Skill Optimization Results}
\label{e}
Table~\ref{tab:skillopt_14categories} shows that SkillCT-guided optimization improves Structural Completeness and Description-Content Consistency across all 14 categories under both GLM-4.5-Air and DeepSeek-V4-Flash. Categories with weaker initial specifications, such as Productivity, generally achieve larger gains than those with stronger baselines, such as Security. This suggests that SkillCT can identify missing procedural details and inconsistencies between Skill descriptions and instructions that can be addressed through refinement. Conciseness changes little, as greater instructional completeness does not necessarily require shorter specifications. While both LLMs show similar improvements in specification-related dimensions, Dynamic and Human-Assisted scores vary more across categories and models. For instance, DeepSeek-V4-Flash scores higher on Intent-Perturbation Stability for Creative Skills, whereas GLM-4.5-Air scores higher for Security Skills, potentially reflecting differences in task interpretation and behavioral evaluation. These results suggest that SkillCT provides useful optimization signals across Skill categories and LLM backbones while capturing differences between specification quality and assessed behavior.

\section{Category-Wise Assessment Agreement}
\label{app:model_agreement}
We measure cross-model agreement using Spearman's $\rho$ for category rankings and MAE for overall score differences. As shown in Table~\ref{tab:category_model_agreement}, the relative rankings and overall scores remain largely consistent across LLMs. For Behavior-Chain Semantic Consistency, category rankings show somewhat greater variation while maintaining broadly similar patterns across models. This difference may arise because models generate different action sequences and vary in judging whether steps are relevant, coherent, or redundant. Such variations can alter category rankings in this dimension without substantially affecting the overall assessment. These findings support the stability of SkillCT for category-wise quality comparisons.

\begin{table}[]
\centering
\caption{Category-level assessment agreement across LLMs on the SkillCT reference benchmark.}
\vspace{-7pt}
\label{tab:category_model_agreement}
\scriptsize
\setlength{\tabcolsep}{3pt}
\renewcommand{\arraystretch}{1.05}
\resizebox{0.8\columnwidth}{!}{%
\begin{tabular}{@{}lccccc@{}}
\toprule
\textbf{Model Pair}
& \multicolumn{2}{c}{\textbf{Overall Score}}
& \multicolumn{3}{c}{\textbf{Dimension ($\rho$)}} \\
\cmidrule(lr){2-3}
\cmidrule(l){4-6}
& $\rho$ & MAE
& \textbf{IPS}
& \textbf{BCSC}
& \textbf{TAG} \\
\midrule
Seed--GLM       & 0.978 & 0.021 & 0.873 & 0.763 & 0.887 \\
Seed--Qwen      & 0.991 & 0.003 & 0.974 & 0.915 & 0.970 \\
Seed--DeepSeek  & 0.990 & 0.028 & 0.976 & 0.974 & 0.998 \\
Seed--GPT-5     & 0.991 & 0.007 & 0.955 & 0.904 & 0.977 \\
GLM--Qwen       & 0.991 & 0.022 & 0.807 & 0.833 & 0.908 \\
GLM--DeepSeek   & 0.988 & 0.009 & 0.866 & 0.793 & 0.893 \\
GLM--GPT-5      & 0.991 & 0.014 & 0.937 & 0.840 & 0.947 \\
Qwen--DeepSeek  & 0.999 & 0.030 & 0.978 & 0.928 & 0.977 \\
Qwen--GPT-5     & 0.991 & 0.009 & 0.926 & 0.929 & 0.960 \\
DeepSeek--GPT-5 & 0.988 & 0.022 & 0.957 & 0.950 & 0.981 \\
\midrule
\textbf{Average}
& \textbf{0.990}
& \textbf{0.016}
& \textbf{0.925}
& \textbf{0.883}
& \textbf{0.950} \\
\bottomrule
\end{tabular}%
}
\parbox{\columnwidth}{%
\scriptsize
\textit{Note.} IPS, BCSC, and TAG denote
Intent-Perturbation Stability,
Behavior-Chain Semantic Consistency,
and Task Abstraction Granularity, respectively.
Seed, GLM, Qwen, and DeepSeek denote
Seed-1.6-Flash, GLM-4.5-Air,
Qwen3.5-Flash, and DeepSeek-V4-Flash, respectively.
}
\vspace{-10pt}
\end{table}

\section{Supplementary Details of the SkillCT Method}
\label{f}
To facilitate understanding and reproducibility, we provide detailed scoring procedures for the SkillCT assessment dimensions.
\subsection{Static Assessment Scoring Details}
\label{app:static_scoring}

\paragraph{Structural Completeness.}
Let $x_i(s)\in\{0,1\}$ indicate whether Skill $s$ satisfies criterion $i$, and let $\alpha_i$ denote its weight. The Structural Completeness score is:
\begin{equation}
S_{\mathrm{SC}}(s)
=
\frac{
\sum_{i=1}^{13}\alpha_i x_i(s)
}{
\sum_{i=1}^{13}\alpha_i
}.
\end{equation}
The criteria and weights are listed in
Table~\ref{tab:structural_details}.

\begin{table}[]
\centering
\caption{Structural Completeness criteria and weights.}
\vspace{-7pt}
\label{tab:structural_details}
\small
\scalebox{0.9}{%
\begin{tabular}{@{}p{0.32\linewidth} p{0.45\linewidth} c@{}}
\toprule
\textbf{Criterion} & \textbf{Condition} & \textbf{Weight} \\
\midrule
Frontmatter & YAML frontmatter is present & 2.0 \\
Name & A Skill name is provided & 1.5 \\
Description & A description longer than 10 characters is provided & 2.5 \\
Description length & Description length is between 10 and 500 characters & 1.0 \\
Instructional body & The instructional body contains more than 50 characters & 2.0 \\
Section structure & At least two second-level sections are present & 1.5 \\
Workflow & Workflow- or procedure-related information is present & 2.0 \\
Code blocks & At least one code block is provided & 1.0 \\
Constraints & Constraint- or limitation-related information is present & 2.0 \\
Error handling & Error- or failure-handling information is present & 1.5 \\
Input-output specification & Input, output, or format information is provided & 1.5 \\
Prerequisites & Prerequisite or dependency information is provided & 1.0 \\
Auxiliary scripts & Supporting scripts are included & 0.5 \\
\bottomrule
\end{tabular}%
}
\end{table}

\paragraph{Conciseness.}
Conciseness is derived from five components: length adequacy, repeated content,
unique-content density, character-level information entropy, and empty
structure.

Let $N_s$ denote the total number of lines in the instructional body of Skill
$s$. The length adequacy score $l(s)$ is defined as

\begin{equation}
l(s)=
\left\{
\begin{array}{ll}
0.3,
& N_s < 10, \\[3pt]

0.5 + 0.5\displaystyle\frac{N_s-10}{20},
& 10 \leq N_s < 30, \\[6pt]

1.0,
& 30 \leq N_s \leq 300, \\[3pt]

1.0 - 0.3\displaystyle\frac{N_s-300}{300},
& 300 < N_s \leq 600, \\[6pt]

0.7 - 0.4\displaystyle\frac{N_s-600}{900},
& 600 < N_s \leq 1500, \\[6pt]

\max\left\{
0.1,\,
0.3 - 0.2\displaystyle\frac{N_s-1500}{3000}
\right\},
& N_s > 1500.
\end{array}
\right.
\end{equation}

To measure repetition, non-empty lines containing at least five characters are
retained. Let $\mathcal{L}_s$ denote this collection and let $m_j$ denote the
number of occurrences of the $j$-th distinct line. The repeated-content ratio
is:
\begin{equation}
r(s)
=
\frac{
\displaystyle\sum_j \max\left(m_j-1,0\right)
}{
|\mathcal{L}_s|
}.
\end{equation}
Let $U_s$ denote the number of unique non-empty lines and $E_s$ the number of
empty lines. Unique-content density $d(s)$ and empty-line ratio $e(s)$ are
computed as:
\begin{equation}
d(s)=\frac{U_s}{N_s},
\qquad
e(s)=\frac{E_s}{N_s}.
\end{equation}
For character-level information diversity, let $\mathcal{A}_s$ denote the set
of characters appearing in the instructional body and let $p_s(a)$ denote the
empirical frequency of character $a\in\mathcal{A}_s$. The normalized entropy
score is:
\begin{equation}
h(s)
=
\min\left\{
1,\,
\frac{
-\displaystyle\sum_{a\in\mathcal{A}_s}
p_s(a)\log_2 p_s(a)
}{5}
\right\}.
\end{equation}
The final Conciseness score is:
\begin{equation}
\begin{aligned}
S_{\mathrm{Con}}(s) ={}& 0.25l(s)
+ 0.20\left(1-r(s)\right)
+ 0.20d(s) \\
&+ 0.20h(s)
+ 0.15\left(1-e(s)\right).
\end{aligned}
\end{equation}

\paragraph{Description-Content Consistency.}
After lowercasing, tokenization, and stop-word removal, let $\mathcal{D}_s$ and $\mathcal{B}_s$ denote the description and body token sets, $\mathcal{G}_s^d$ and $\mathcal{G}_s^b$ the bigram sets constructed from their ordered token sequences, and $\mathcal{T}_s$ the section-title token set.
Keyword coverage $\kappa(s)$, bigram coverage $\beta(s)$, section-title
coverage $\sigma(s)$, and token-level Jaccard similarity $\zeta(s)$ are
computed as:
\begin{equation}
\begin{array}{ll}
\displaystyle
\kappa(s)
=
\frac{
|\mathcal{D}_s\cap\mathcal{B}_s|
}{
|\mathcal{D}_s|
},
&
\displaystyle
\beta(s)
=
\frac{
|\mathcal{G}_s^{d}\cap\mathcal{G}_s^{b}|
}{
|\mathcal{G}_s^{d}|
},
\\[14pt]

\displaystyle
\sigma(s)
=
\frac{
|\mathcal{D}_s\cap\mathcal{T}_s|
}{
|\mathcal{D}_s|
},
&
\displaystyle
\zeta(s)
=
\frac{
|\mathcal{D}_s\cap\mathcal{B}_s|
}{
|\mathcal{D}_s\cup\mathcal{B}_s|
}.
\end{array}
\end{equation}

Let $\mathbf{v}_d$ and $\mathbf{v}_b$ denote the TF-IDF vectors of the description and body, using length-normalized term frequencies and $\mathrm{idf}(w)=\log(2/\mathrm{df}(w))+1$, where $\mathrm{df}(w)$ is computed over the description-body pair. Their cosine similarity is
\begin{equation}
\rho(s)
=
\frac{
\mathbf{v}_d^{\top}\mathbf{v}_b
}{
\|\mathbf{v}_d\|_2
\|\mathbf{v}_b\|_2
}.
\end{equation}

The final Description-Content Consistency score is
\begin{equation}
\begin{aligned}
S_{\mathrm{DC}}(s) ={}& 0.30\kappa(s)
+ 0.30\rho(s)
+ 0.20\beta(s) \\
&+ 0.10\sigma(s)
+ 0.10\min\left\{10\zeta(s),1\right\}.
\end{aligned}
\end{equation}
If the Skill description is absent or contains no valid tokens after
preprocessing, $S_{\mathrm{DC}}(s)$ is set to zero.

\subsection{Dynamic Assessment Scoring Details}
\label{app:dynamic_scoring}

\paragraph{Cross-Model Robustness.}
For each Skill, we submit the same task request to four backbone models. Let $M$ denote the number of valid responses. For the $i$th response, let $a_i$ denote its applicability judgment, $L_i$ the number of steps in its action plan, and $T_i$ the set of selected tools or commands. We measure agreement across models using applicability consistency, plan length consistency, and tool consistency:
\begin{equation}
\begin{aligned}
A_{\mathrm{app}}(s)
&=
\frac{1}{M}
\max_{c\in\mathcal{C}}
\sum_{i=1}^{M}\mathbf{1}(a_i=c),\\
A_{\mathrm{plan}}(s)
&=
\max\left\{0,\,1-\frac{\sigma_L}{\mu_L}\right\},\\
A_{\mathrm{tool}}(s)
&=
\frac{1}{\binom{M}{2}}
\sum_{1\leq i<j\leq M}J(T_i,T_j),
\end{aligned}
\end{equation}
where $\mathcal{C}$ is the set of possible applicability judgments, and $\mu_L$ and $\sigma_L$ are the mean and standard deviation of the plan lengths, respectively. When all returned plans are empty, $A_{\mathrm{plan}}(s)$ is set to $0.5$. The Jaccard similarity $J(T_i,T_j)$ is set to $1$ when both sets are empty and is computed normally otherwise.

The final Cross-Model Robustness score combines the three components:
\begin{equation}
\begin{aligned}
S_{\mathrm{CMR}}(s) ={}&
0.35A_{\mathrm{app}}(s)
+0.35A_{\mathrm{plan}}(s)\\
&+0.30A_{\mathrm{tool}}(s).
\end{aligned}
\end{equation}




\paragraph{Intent-Perturbation Stability.}
Five intent-preserving request variants are generated using synonym
replacement, sentence reordering, insertion of irrelevant context, concise
reformulation, and formal reformulation. Each successful variant is compared
with the execution of the original request.

For perturbation $k$, applicability consistency is $1$ when its applicability
judgment matches that of the original request and $0$ otherwise. If $L_0$ and
$L_k$ denote the numbers of plan steps for the original and perturbed
requests, plan consistency is
\begin{equation}
B_{\mathrm{plan}}^{k}(s)
=
1-
\frac{|L_0-L_k|}
{\max(L_0,L_k)},
\end{equation}

with a value of $1$ when both plans are empty. Tool consistency is the
Jaccard similarity between the corresponding tool or command sets. The score
for perturbation $k$ is
\begin{equation}
R_k(s)
=
0.40B_{\mathrm{app}}^{k}(s)
+
0.30B_{\mathrm{plan}}^{k}(s)
+
0.30B_{\mathrm{tool}}^{k}(s).
\end{equation}

The final Intent-Perturbation Stability score is averaged over the $K_{\mathrm{valid}}$ successfully evaluated variants:
\begin{equation}
S_{\mathrm{IPS}}(s)
=
\frac{1}{K_{\mathrm{valid}}}
\sum_{k=1}^{K_{\mathrm{valid}}} R_k(s).
\end{equation}
If no variant returns a valid result, the score is set to zero.

\paragraph{Capability-Boundary Handling.}
SkillCT evaluates four request types: unrelated, adversarial,
adjacent-domain, and excessive-scope requests. Each response is scored based on its applicability judgment, explanation, confidence, and proposed plan. Table~\ref{tab:boundary_scoring} summarizes the scoring rules.
\begin{table}[t]
\centering
\caption{Scoring rules for Capability-Boundary Handling.}
\label{tab:boundary_scoring}
\small
\setlength{\tabcolsep}{4pt}
\resizebox{\columnwidth}{!}{%
\begin{tabular}{@{}llc@{}}
\toprule
\textbf{Request Type} & \textbf{Response} & \textbf{Score} \\
\midrule

\multirow{3}{*}{Unrelated}
& Identified as unsupported with an explanation & 1.0 \\
& Identified as unsupported without an explanation & 0.8 \\
& Treated as supported & 0.0 \\
\midrule

\multirow{3}{*}{Adversarial}
& Identified as unsupported with an explanation & 1.0 \\
& Identified as unsupported without an explanation & 0.8 \\
& Treated as supported & 0.0 \\
\midrule

\multirow{5}{*}{\shortstack[l]{Adjacent\\domain}}
& Identified as unsupported with an explanation & 1.0 \\
& Identified as unsupported without an explanation & 0.8 \\
& Treated as supported with low confidence & 0.5 \\
& High-confidence acceptance with a proposed plan & 0.3 \\
& Other partially supported cases & 0.5 \\
\midrule

\multirow{4}{*}{\shortstack[l]{Excessive\\scope}}
& Identified as unsupported with an explanation & 1.0 \\
& Identified as unsupported without an explanation & 0.8 \\
& High-confidence acceptance & 0.2 \\
& Other partially supported cases & 0.5 \\
\bottomrule
\end{tabular}%
}
\end{table}
If a test does not return a valid response, it receives a neutral score of $0.5$. The final Capability-Boundary Handling score is the average across the four request types.

\begin{table}[t]
\centering
\caption{Normalization settings for Efficiency.}
\label{tab:efficiency_ranges}
\resizebox{0.9\linewidth}{!}{%
\begin{tabular}{@{}lccc@{}}
\toprule
\textbf{Quantity} & \shortstack{\textbf{Operating}\\\textbf{Interval}}
& \shortstack{\textbf{Upper}\\\textbf{Threshold}} & \textbf{Weight} \\
\midrule
Prompt tokens      & 500--2000     & 8000     & 0.30 \\
Plan steps         & 3--7          & 20       & 0.25 \\
Response latency   & 500--3000 ms  & 15000 ms & 0.20 \\
Tools or commands  & 1--5          & 15       & 0.25 \\
\bottomrule
\end{tabular}%
}
\end{table}

\paragraph{Efficiency.}
Efficiency is computed from prompt-token consumption, the number of planned
steps, response latency, and the number of selected tools or commands. Each
quantity is converted to a normalized score using the same piecewise function.

For an observed value $x$, an operating interval $[l,u]$, and an upper
threshold $m$, the normalization function is
\begin{equation}
g(x;l,u,m)
=
\left\{
\begin{array}{ll}
0.6,
& x \leq 0, \\[3pt]

0.6 + 0.4\displaystyle\frac{x}{l},
& 0 < x < l, \\[6pt]

1.0,
& l \leq x \leq u, \\[3pt]

\max\left\{
0.2,\,
1.0-0.8\displaystyle\frac{x-u}{m-u}
\right\},
& u < x \leq m, \\[6pt]

0.1,
& x > m.
\end{array}
\right.
\end{equation}
The parameters used for the four quantities are listed in
Table~\ref{tab:efficiency_ranges}.

Let $E_{\mathrm{tok}}(s)$, $E_{\mathrm{step}}(s)$,
$E_{\mathrm{lat}}(s)$, and $E_{\mathrm{tool}}(s)$ denote the four normalized
scores. The final Efficiency score is:
\begin{equation}
S_{\mathrm{Eff}}(s)
=
0.30E_{\mathrm{tok}}(s)
+
0.25E_{\mathrm{step}}(s)
+
0.20E_{\mathrm{lat}}(s)
+
0.25E_{\mathrm{tool}}(s).
\end{equation}

\subsection{Human-Assisted Assessment Scoring Details}
\label{app:human_scoring}
Tables~\ref{tab:behavior_chain_rubric} and~\ref{tab:granularity_rubric} provide the detailed scoring rubrics for the two Human-Assisted Assessment dimensions.
\begin{table}[h]
\centering
\caption{Scoring rubric for Behavior-Chain Semantic Consistency.}
\label{tab:behavior_chain_rubric}
\small
\begin{tabular}{@{}cp{0.9\linewidth}@{}}
\toprule
\textbf{Score} & \textbf{Criterion} \\
\midrule
1.00 & All actions directly support the declared objective, with no irrelevant or redundant operations. \\
0.75 & Most actions support the objective, with only one or two peripheral actions. \\
0.50 & Only part of the behavior chain supports the objective, with clear irrelevant or redundant operations. \\
0.25 & Most actions are weakly related or unrelated to the declared objective, resulting in a poorly aligned behavior chain. \\
0.00 & The behavior chain is almost entirely unrelated to the declared objective. \\
\bottomrule
\end{tabular}
\end{table}

\begin{table}[h]
\centering
\caption{Scoring rubric for Task Abstraction Granularity.}
\label{tab:granularity_rubric}
\small
\begin{tabular}{@{}cp{0.82\linewidth}@{}}
\toprule
\textbf{Score} & \textbf{Criterion} \\
\midrule
1.00 & The Skill represents a complete and meaningful unit of work with a clear and self-contained responsibility. \\
0.75 & The task scope is generally appropriate but is slightly narrower or broader than the intended unit of work. \\
0.50 & The responsibility is noticeably too narrow or too broad. \\
0.25 & The task scope is severely misaligned and would substantially benefit from being split or combined with other functionality. \\
0.00 & The Skill either reduces to an isolated operation with little additional procedure or attempts to cover an entire domain without a coherent task scope. \\
\bottomrule
\end{tabular}
\end{table}






\end{document}